%% file: emnlp2018.tex
\documentclass[11pt,a4paper]{article}
\usepackage[hyperref]{emnlp2018}
\usepackage{times}
\usepackage{latexsym}
\usepackage{graphicx}
\usepackage{amsmath}
\usepackage{amssymb, amsfonts, ulem, xspace, color}
\usepackage{array}
\usepackage[T1]{fontenc}
\usepackage[utf8]{inputenc}
\usepackage[russian,english]{babel}
\usepackage{CJKutf8}
\usepackage{booktabs}
\usepackage{subcaption}
\usepackage{url}

\DeclareMathOperator*{\argmin}{argmin}

\newcommand{\red}[1]{\textcolor{red}{\textbf{{#1}}}}

\aclfinalcopy 
\def\aclpaperid{1605} 

\title{XNLI: Evaluating Cross-lingual Sentence Representations}

\author{Alexis Conneau, Guillaume Lample, Ruty Rinott, Holger Schwenk, Ves Stoyanov  \\
  Facebook AI \\
  {\tt \normalsize aconneau,glample,ruty,schwenk,ves@fb.com} 
  \AND
  Adina Williams, Samuel R. Bowman \\
  New York University \\
  {\tt \normalsize adinawilliams,bowman@nyu.edu} \\
  }

\date{}

\hypersetup{draft}
\begin{document}
\maketitle
\input{content/tables}

\insertexamples
\newcommand{\nlangs}{15\xspace}

\begin{abstract}

State-of-the-art natural language processing systems rely on supervision in the form of annotated data to learn competent models. These models are generally trained on data in a single language (usually English), and cannot be directly used beyond that language.
Since collecting data in every language is not realistic, there has been a growing interest in cross-lingual language understanding (XLU) and low-resource cross-language transfer.
In this work, we construct an evaluation set for XLU by extending the development and test sets of the Multi-Genre Natural Language Inference Corpus (MultiNLI) 
to \nlangs languages, including low-resource languages such as Swahili and Urdu. We hope that our dataset, dubbed XNLI, will catalyze research in cross-lingual sentence understanding by providing an informative standard evaluation task.
In addition, we provide several baselines for multilingual sentence understanding, including two based on machine translation systems, and two that use parallel data to train aligned multilingual bag-of-words and LSTM encoders.
We find that XNLI represents a practical and challenging evaluation suite, and that directly translating the test data yields the best performance among available baselines.

\end{abstract}

\section{Introduction}

Contemporary natural language processing systems typically rely on annotated data to learn how to perform a task (e.g., classification, sequence tagging, natural language inference).
Most commonly the available training data is in a single language (e.g., English or Chinese) and the resulting system can perform the task only in the training language. In practice, however, systems used in major international products 
need to handle inputs in many languages. In these settings, it is nearly impossible to annotate data in  all languages that a system might encounter during operation. 

A scalable way to build multilingual systems is through cross-lingual language understanding (XLU), in which a system is trained primarily on data in one language and evaluated on data in others.
While XLU shows promising results for tasks such as cross-lingual document classification \cite{Klementiev:2012:coling_reuters,Schwenk:2018:lrec_mldoc}, there are very few, if any, XLU benchmarks for more difficult language understanding tasks like natural language inference.
Large-scale natural language inference (NLI), also known as recognizing textual entailment (RTE), has emerged as a practical test bed for work on sentence understanding. In NLI, a system is tasked with reading two sentences and determining whether one entails the other, contradicts it, or neither (\textit{neutral}). Recent crowdsourced annotation efforts have yielded datasets for NLI in English \cite{bowman2015large,multinli:2017} with nearly a million examples, and these have been widely used to evaluate neural network architectures and training strategies \cite{rocktaschel2015reasoning, gong2017natural, peters2018deep, wang2018glue}, as well as to train effective, reusable sentence representations  \cite{conneau2017supervised,subramanian2018learning, cer2018universal,conneau2018craminto}.

In this work, we introduce a benchmark that we call the Cross-lingual Natural Language Inference corpus, or XNLI, by extending these NLI corpora to \nlangs languages.  
XNLI consists of 7500 human-annotated development and test examples in NLI three-way classification format in 
English, French, Spanish, German, Greek, Bulgarian, Russian, Turkish, Arabic, Vietnamese, Thai, Chinese, Hindi, Swahili and Urdu, making a total of 112,500 annotated pairs. These languages span several language families, and with the inclusion of Swahili and Urdu, include two lower-resource languages as well. 

Because of its focus on development and test data, this corpus is designed to evaluate \textit{cross-lingual} sentence understanding, where models have to be trained in one language and tested in different ones.

We evaluate several approaches to cross-lingual learning of natural language inference that leverage parallel data from publicly available corpora at training time. We show that parallel data can help align sentence encoders in multiple languages such that a classifier trained with English NLI data can correctly classify pairs of sentences in other languages. While outperformed by our machine translation baselines, we show that this alignment mechanism gives very competitive results.

A second practical use of XNLI is the evaluation of pretrained general-purpose language-universal sentence encoders.
We hope that this benchmark will help the research community build multilingual text embedding spaces. Such embeddings spaces will facilitate the creation of multilingual systems that can transfer across languages with little or no extra supervision.

The paper is organized as follows: We next survey the related literature on cross-lingual language understanding. We then describe our data collection methods and the resulting corpus in Section~\ref{SectCorpus}. We describe our baselines in Section~\ref{SectBaselines}, and finally present and discuss results in Section~\ref{SectExps}.

\section{Related Work}

\paragraph{Multilingual Word Embeddings}
Much of the work on multilinguality in language understanding has been at the word level. Several approaches have been proposed to learn cross-lingual word representations, i.e., word representations where translations are close in the embedding space. Many of these methods require some form of super\-vision (typically in the form of a small bi\-lingual lexicon) to align two sets of source and target embeddings to the same space~\cite{mikolov2013exploiting,Kocisky:2014:acl_biword,Faruqui:2014:eacl,ammar2016massively}. More recent studies have showed that cross-lingual word embeddings can be generated with no supervision whatsoever~\cite{Artetxe:2017:acl_mlword,Conneau:2018:iclr_muse}.

\paragraph{Sentence Representation Learning}
Many approaches have been proposed to extend word embeddings to sentence or paragraph representations~\cite{le2014distributed,wieting2015towards,arora2016asimple}. The most straightforward way to generate sentence embeddings is to consider an average or weighted average of word representations, usually referred to as continuous bag-of-words (CBOW). Although na\"ive, this method often provides a strong baseline. More sophisticated approaches---such as the un\-supervised SkipThought model of~\citet{kiros2015skip} that extends the skip-gram model of~\citet{mikolov2013distributed} to the sentence level---have been proposed to capture syntactic and semantic depend\-encies inside sentence representations. While these fixed-size sentence embedding methods have been outperformed by their supervised counterparts~\cite{conneau2017supervised,subramanian2018learning}, some recent developments have shown that pretrained language models can also transfer very well, either when the hidden states of the model are used as contextualized word vectors~\cite{peters2018deep}, or when the full model is fine-tuned on transfer tasks~\cite{radford2018improving,howard2018fine}.

\paragraph{Multilingual Sentence Representations}
There has been some effort on developing multilingual sentence embeddings.
For example, \citet{Chandar:2013:multlin_deep} train bi\-lingual auto\-encoders with the objective of minimizing reconstruction error between two languages.  \citet{Schwenk:2017:repl4nlp} and \citet{epsana:2017:ieee_embed_mine} jointly train a sequence-to-sequence MT system on multiple languages to learn a shared multilingual sentence embedding space. \citet{Hermann:2014:acl_multling} propose a compositional vector model involving unigrams and bigrams to learn document level representations.
\citet{Pham:2015:multling} directly train embedding representations for sentences with no attempt at compositionality. \citet{Zhou:2016:acl_crossling} learn bilingual document representations by minimizing the Euclidean distance between document representations and their translations.

\paragraph{Cross-lingual Evaluation Benchmarks}
The lack of evaluation benchmark has hindered the development of such multilingual representations. Most previous approaches use the Reuters cross-lingual document classification corpus~\citet{Klementiev:2012:coling_reuters} for evaluation. However, the classification in this corpus is done at document level, and, as there are many ways to aggregate sentence embeddings, the comparison between different sentence embeddings is difficult. Moreover, the distribution of classes in the Reuters corpus is highly unbalanced, and the dataset does not provide a development set in the target language, further complicating experimental comparisons.

In addition to the Reuters corpus, \citet{cer2017semeval} propose sentence-level multilingual training and evaluation datasets for semantic textual similarity in four languages. There have also been efforts to build multilingual RTE datasets, either through translating English data \cite{Mehdad:2011}, or annotating sentences from a parallel corpora \cite{Negri:2011}. More recently, \citet{agic2017baselines} provide a corpus, that is very complementary to our work, of human translations for 1332 pairs of the SNLI data into Arabic, French, Russian, and Spanish. Among all these benchmarks, XNLI is the first large-scale corpus for evaluating sentence-level representations on that many languages.

In practice, cross-lingual sentence understanding goes beyond translation. For instance, \citet{Mohammad:2016} analyze the differences in human sentiment annotations of Arabic sentences and their English translations, and conclude that most of them come from cultural differences. Similarly, \citet{Smith:2016} show that most of the degradation in performance when applying a classification model trained in English to Spanish data translated to English is due to cultural differences.
One of the limitations of the XNLI corpus is that it does not capture these differences, since it was obtained by translation. We see the XNLI evaluation as a necessary step for multilingual NLP before tackling the even more complex problem of domain-adaptation that occurs when handling this the change in style from one language to another.

\section{The XNLI Corpus}
\label{SectCorpus}

Because the test portion of the Multi-Genre NLI data was kept private, the Cross-lingual NLI Corpus (XNLI) is based on new English NLI data. To collect the core English portion, we follow precisely the same crowdsourcing-based procedure used for the existing Multi-Genre NLI corpus, and collect and validate 750 new examples from each of the ten text sources used in that corpus for a total of 7500 examples. With that portion in place, we create the full XNLI corpus by employing professional translators to translate it into our ten target languages. This section describes this process and the resulting corpus. 

Translating, rather than generating new hypothesis sentences in each language separately, has multiple advantages. First, it ensures that the data distributions are maximally similar across languages. As speakers of different languages may have slightly different intuitions about how to fill in the supplied prompt, this allows us to avoid adding this unwanted degree of freedom. Second, it allows us to use the same trusted pool of workers as was used prior NLI crowdsourcing efforts, without the need for training a new pool of workers in each language. Third, for any premise, this process allows us to have a corresponding hypothesis in any language. XNLI can thus potentially be used to evaluate whether an Arabic or Urdu premise is entailed with a Bulgarian or French hypothesis etc. This results in more than 1.5M combinations of hypothesis and premises. Note that we do not consider that use case in this work.

This translation approach carries with it the risk that the semantic relations between the two sentences in each pair might not be reliably preserved in translation, as \citet{Mohammad:2016} observed for sentiment. We investigate this potential issue in our corpus and find that, while it does occur, it only concerns a negligible number of sentences (see Section~\ref{sec:analysis}).

\subsection{Data Collection}

\paragraph{The English Corpus}

Our collection procedure for the English portion of the XNLI corpus follows the same procedure as the MultiNLI corpus. We sample 250 sentences from each of the ten sources that were used in that corpus, ensuring that none of those selected sentences overlap with the distributed corpus. Nine of the ten text sources are drawn from the second release of the Open American National Corpus\footnote{\url{http://www.anc.org/}}: \textit{Face-To-Face}, \textit{Telephone}, \textit{Government}, \textit{9/11}, \textit{Letters}, \textit{Oxford University Press} (\textit{OUP}), \textit{Slate}, \textit{Verbatim}, and \textit{Government}. The tenth, \textit{Fiction}, is drawn from the novel {\it Captain Blood} \cite{CaptainBlood}. We refer the reader to \citet{multinli:2017} for more details on each genre.

Given these sentences, we ask the same MultiNLI worker pool from a crowdsourcing platform to produce three hypotheses for each premise, one for each possible label.

We present premise sentences to workers using the same templates as were used in MultiNLI. We also follow that work in pursuing a second validation phase of data collection in which each pair of sentences is relabeled by four other workers. For each validated sentence pair, we assign a gold label representing a majority vote between the initial label assigned to the pair by the original
annotator, and the four additional labels assigned
by validation annotators. We obtained a three-vote consensus for 93\% of the data. In our experiments, we kept the 7\% additional ones, but we mark these ones with a special label '-'. 

\insertstats

\paragraph{Translating the Corpus}

Finally, we hire translators to translate the resulting sentences into \nlangs languages using the {\it One Hour Translation} platform. We translate the premises and hypotheses separately, to ensure that no context is added to the hypothesis that was not there originally, and simply copy the labels from the English source text. Some development examples are shown in Table~\ref{tab:examples}. 

\subsection{The Resulting Corpus}
\label{sec:analysis}

One main concern in studying the resulting corpus is to determine whether the gold label for some of the sentence pairs changes as a result of information added or removed in the translation process. 

Investigating the data manually, we find an example in the Chinese translation where an entailment relation becomes a contradictory relation, while the entailment is preserved in other languages. Specifically, the term {\it upright} which was used in English as entailment of {\it standing}, was translated into Chinese as {\it sitting upright} thus creating a contradiction. However, the difficulty of finding such an example in the data suggests its rarity. 

To quantify this observation, we recruit two bilingual annotators to re-annotate 100 examples each in both English and French following our standard validation procedure. The examples are drawn from two non-overlapping random subsets of the development data to prevent the annotators from seeing the source English text for any translated text they annotate. With no training or burn-in period, these annotators recover the English consensus label 85\% of the time on the original English data and 83\% of the time on the translated French, suggesting that the overall semantic relationship between the two languages has been preserved. As most sentences are relatively easy to translate, in particular the hypotheses generated by the workers, there seems to be little ambiguity added by the translator. 

More broadly, we find that the resulting corpus has similar properties to the MultiNLI corpus. For all languages, on average, the premises are twice as long as the hypotheses (See Table~\ref{tab:stats_all}). The top hypothesis words indicative of the class label -- scored using the mutual information between each word and class in the corpus -- are similar across languages, and overlap those of the MultiNLI corpus \cite{gururangan2018annotation}. For example, a translation of at least one of the words {\it no}, {\it not} or {\it never} is among the top two cues for contradiction in all languages.

As in the original MultiNLI corpus, we expect that cues like these \citep[`artifacts', in Gururangan's terms, also observed by ][]{Poliak2018,TSUCHIYA18.786} allow a baseline system to achieve better-than-random accuracy with access only to the premise sentences. We accept this as an unavoidable property of the NLI task over naturalistic sentence pairs, and see no reason to expect that this baseline would achieve better accuracy than the relatively poor 53\% seen in \citet{gururangan2018annotation}.

The current version of the corpus is freely available\footnote{\scriptsize\url{https://s3.amazonaws.com/xnli/XNLI-1.0.zip}}\footnote{\scriptsize\url{https://s3.amazonaws.com/xnli/XNLI-MT-1.0.zip}} for typical machine learning uses, and may be modified and redistributed. The majority of the
corpus sentences are released under the OANC's license which allows all content to be freely used, modified, and shared under permissive terms. The data in the \textit{Fiction} genre from {\it Captain Blood} are in the public domain in the United States (but may be licensed differently elsewhere).


\insertalignment

\section{Cross-Lingual NLI}
\label{SectBaselines}

In this section we present results with XLU systems that can serve as baselines for future work.

\subsection{Translation-Based Approaches}

The most straightforward techniques for XLU rely on translation systems. There are two natural ways to use a translation system:
\textsc{translate train}, where the training data is translated into each target language to provide data to train each classifier, and \textsc{translate test}, where a translation system is used at test time to translate input sentences to the training language. These two methods provide strong baselines, but both present practical challenges. The former requires training and maintaining as many classifiers as there are languages, while the latter relies on computationally-intensive translation at test time. Both approaches are limited by the quality of the translation system, which itself varies with the quantity of available training data and the similarity of the language pair involved.

\subsection{Multilingual Sentence Encoders}
\label{SectMLenc}

An alternative to translation is to rely on language-universal embeddings of text and build multilingual classifiers on top of these representations. If an encoder produces an embedding of an English sentence close to the embedding of its translation in another language, then a classifier learned on top of English sentence embeddings will be able to classify sentences from different languages without needing a translation system at inference time.

We evaluate two types of cross-lingual sentence encoders: (i) pretrained universal multilingual sentence embeddings based on the average of word embeddings (\textsc{x-cbow}), (ii) bidirectional-LSTM~\cite{hochreiter1997long} sentence encoders trained on the MultiNLI training data (\textsc{x-bilstm}). The former evaluates transfer learning while the latter evaluates NLI-specific encoders trained on in-domain data. Both approaches use the same alignment loss for aligning sentence embedding spaces from multiple languages which is present below. We consider two ways of extracting feature vectors from the BiLSTM: either using the initial and final hidden states \cite{sutskever2014sequence}, or using the element-wise max over all states \citep{collobert2008unified}.

The first approach is commonly used as a strong baseline for monolingual sentence embeddings \cite{arora2016asimple, conneau2018senteval, Gouws:2015:icml_bilbowa}. Concretely, we consider the English fastText word embedding space as being fixed, and fine-tune embeddings in other languages so that the average of the word vectors in a sentence is close to the average of the word vectors in its English translation. The second approach consists in learning an English sentence encoder on the MultiNLI training data along with an encoder on the target language, with the objective that the representations of two translations are nearby in the embedding space. In both approaches, an English encoder is fixed, and we train target language encoders to match the output of this encoder. This allows us to build sentence representations that belong to the same space. Joint training of encoders and parameter sharing are also promising directions to improve and simplify the alignment of sentence embedding spaces. We leave this for future work.

In all experiments, we consider encoders that output a vector of fixed size as a sentence representation. While previous work shows that performance on the NLI task can be improved by using cross-sentence attention between the premise and hypothesis \citep{rocktaschel2015reasoning,gong2017natural}, we focus on methods with fixed-size sentence embeddings.

\insertwordtranslationtable

\subsubsection{Aligning Word Embeddings}
\label{mul-word-emb}

Multilingual word embeddings are an efficient way to transfer knowledge from one language to another. For instance, \citet{zhang2016ten} show that cross-lingual embeddings can be used to extend an English part-of-speech tagger to the cross-lingual setting, and \citet{xiao2014distributed} achieve similar results in dependency parsing.
Cross-lingual embeddings also provide an efficient mechanism to bootstrap neural machine translation (NMT) systems for low-resource language pairs, which is critical in the case of un\-supervised machine translation~\cite{unsupNMTlample,unsupNMTartetxe,lample2018phrase}. In that case, the use cross-lingual embeddings directly helps the alignment of sentence-level encoders.
Cross-lingual embeddings can be generated efficiently using a very small amount of super\-vision. By using a small parallel dictionary with $n=5000$ word pairs, it is possible to learn a linear mapping to minimize
\begin{equation*}
\label{eq:procrustes}
W^{\star} = \underset{W \in O_d(\mathbb{R})}{\argmin} \Vert W X - Y \Vert_\mathrm{F} = UV^T,
\end{equation*}
where $d$ is the dimension of the embeddings, and $X$ and $Y$ are two matrices of shape $(d, n)$ that correspond to the aligned word embeddings that appear in the parallel dictionary, $O_d(\mathbb{R})$ is the group of orthogonal matrices of dimension $d$, and $U$ and $V$ are obtained from the singular value decomposition (SVD) of $YX^T$: $U\Sigma V^T = \text{SVD}(YX^T)$. \citet{xing2015normalized} show that enforcing the ortho\-gonality constraint on the linear mapping leads to better results on the word translation task.

In this paper, we pretrain our embeddings using the common-crawl word embeddings \cite{grave2018learning} aligned with the MUSE library of~\citet{Conneau:2018:iclr_muse}.

\subsubsection{Universal Multilingual Sentence Embeddings}
Most of the successful recent approaches for learning universal sentence representations have relied on English \cite{kiros2015skip,arora2016asimple,conneau2017supervised,subramanian2018learning,cer2018universal}. While notable recent approaches  have considered building a shared sentence encoder for multiple languages using publicly available parallel corpora \cite{johnson2016google,Schwenk:2017:repl4nlp,epsana:2017:ieee_embed_mine}, the lack of a large-scale, sentence-level semantic  evaluation has limited their adoption by the community. In particular, these methods do not cover the scale of languages considered in XNLI, and are limited to high-resource languages. As a baseline for the evaluation of pretrained multilingual sentence representations in the \nlangs languages of XNLI, we consider state-of-the-art common-crawl embeddings with a CBOW encoder. 
Our approach, dubbed \textsc{x-cbow}, consists in fixing the English pretrained word embeddings, and fine-tuning the target (e.g., French) word embeddings so that the CBOW representations of two translations are close in embedding space.
In that case, we consider our multilingual sentence embeddings as being pretrained and only learn a classifier on top of them to evaluate their quality, similar to so-called ``transfer'' tasks in \cite{kiros2015skip, conneau2017supervised} but in the multilingual setting.

\subsubsection{Aligning Sentence Embeddings}
\label{sec:alignment}

Training for similarity of source and target sentences in an embedding space is conceptually and computationally simpler than generating a translation in the target language from a source sentence.
We propose a method for training for cross-lingual similarity and evaluate approaches based on the simpler task of aligning sentence representations. Under our objective, the embeddings of two parallel sentences need not be identical, but only close enough in the embedding space that the decision boundary of the English classifier captures the similarity.

We propose a simple alignment loss function to align the embedding spaces of two different languages. Specifically, we train an English encoder on NLI, and train a target encoder by minimizing the loss:

\vspace{-0.23cm}
{\small$$\mathcal{L}_{\text{align}}(x,y) = dist(x, y) - \lambda (dist(x_c, y) + dist(x, y_c))$$}
\vspace{-0.5cm}

where $(x,y)$ corresponds to the source and target sentence embeddings, $(x_c, y_c)$ is a contrastive term (i.e. negative sampling), $\lambda$ controls the weight of the negative examples in the loss. For the distance measure, we use the L2 norm $dist(x,y) = \Vert x-y \Vert_2$. A ranking loss~\cite{weston2011wsabie} of the form

\vspace{-0.23cm}
{\small
\begin{eqnarray*}
\mathcal{L}_{\text{rank}}(x,y) = \max (0, \alpha - dist(x, y_c) + dist(x, y)) \ + \\
\max (0, \alpha - dist(x_c, y) + dist(x, y))
\end{eqnarray*}
}
\vspace{-0.5cm}

that pushes the sentence embeddings of a translation pair to be closer than the ones of negative pairs leads to very poor results in this particular case. As opposed to $\mathcal{L}_{\text{align}}$, $\mathcal{L}_{\text{rank}}$ does not force the embeddings of sentence pairs to be close enough so that the shared classifier can understand that these sentences have the same meaning.

We use $\mathcal{L}_{\text{align}}$ in the cross-lingual embeddings baselines \textsc{x-cbow}, \textsc{x-bilstm-last} and \textsc{x-bilstm-max}. For \textsc{x-cbow}, the encoder is pretrained and not fine-tuned on NLI (transfer-learning), while the English X-BiLSTMs are trained on the MultiNLI training set (in-domain). For the three methods, the English encoder and classifier are then fixed. Each of the 14 other languages have their own encoders with same architecture. These encoders are trained to "copy" the English encoder using the $\mathcal{L}_{\text{align}}$ loss and the parallel data described in section ~\ref{SectDataParallel}. Our sentence embedding alignment approach is illustrated in Figure~\ref{fig:alignment}.

We only back-propagate through the target encoder when optimizing $\mathcal{L}_{\text{align}}$ such that all 14 encoders live in the same English embedding space. In these experiments, we initialize lookup tables of the LSTMs with pretrained cross-lingual embeddings discussed in Section~\ref{mul-word-emb}.

\section{Experiments and Results}
\label{SectExps}

\insertXNLIresults

\subsection{Training details}
We use internal translation systems to translate data between English and the 10 other languages. For {\sc translate test} (see Table~\ref{tab:results_all}), we translate each test set into English, while for the {\sc translate train}, we translate the English training data of MultiNLI\footnote{To allow replication of results, we share the MT translations of XNLI training and test sets.}.
To give an idea of the translation quality, we give BLEU scores of the automatic translation from the foreign language into English of the XNLI test set in Table~\ref{tab:bleu}. We use the MOSES tokenizer for most languages, falling back on the default English tokenizer when necessary. We use the Stanford segmenter for Chinese \cite{chang2008optimizing}, and the \textit{pythainlp} package for Thai.

We use pretrained 300D aligned word embeddings for both \textsc{x-cbow} and \textsc{x-bilstm} and only consider the most 500,000 frequent words in the dictionary, which generally covers more than 98\% of the words found in XNLI corpora.  We set the number of hidden units of the BiLSTMs to 512, and use the Adam optimizer \cite{kingma2014adam} with default parameters. As in \cite{conneau2017supervised}, the classifier receives a vector $[u,v,\vert u-v \vert, u*v]$, where $u$ and $v$ are the embeddings of the premise and hypothesis provided by the shared encoder, and $*$ corresponds to the element-wise multiplication (see Figure~\ref{fig:alignment}). For the alignment loss, setting $\lambda$ to $0.25$ worked best in our experiments, and we found that the trade-off between the importance of the positive and the negative pairs was particularly important (see Table~\ref{tab:ablation}). We sample negatives randomly. When fitting the target BiLSTM encoder to the English encoder, we fine-tune the lookup table associated to the target encoder, but keep the source word embeddings fixed. The classifier is a feed-forward neural network with one hidden layer of 128 hidden units, regularized with dropout~\cite{srivastava2014dropout} at a rate of $0.1$. For X-BiLSTMs, we perform model selection on the XNLI validation set in each target language. For X-CBOW, we keep a validation set of parallel sentences to evaluate our alignment loss. The alignment loss requires a parallel dataset of sentences for each pair of languages, which we describe next.

\subsection{Parallel Datasets}
\label{SectDataParallel}
We use publicly available parallel datasets to learn the alignment between English and target encoders. For French, Spanish, Russian,  Arabic and Chinese, we use the United Nation corpora \cite{ziemski2016united}, for German, Greek and Bulgarian, the Europarl corpora~\cite{koehn2005europarl}, for Turkish, Vietnamese and Thai, the OpenSubtitles 2018 corpus~\cite{TIEDEMANN12.463}, and for Hindi, the IIT Bombay corpus \cite{kunchukuttan2018iit}.  For all the above language pairs, we were able to gather more than 500,000 parallel sentences, and we set the maximum number of parallel sentences to 2 million.
For the lower-resource languages Urdu and Swahili, the number of parallel sentences is an order of magnitude smaller than for the other languages we consider. For Urdu, we used the Bible and Quran transcriptions~\cite{TIEDEMANN12.463}, the OpenSubtitles 2016~\cite{lison2016} and 2018 corpora and LDC2010T21, LDC2010T23 LDC corpora, and obtained a total of 64k parallel sentences. For Swahili, we were only able to gather 42k sentences using the Global Voices corpus and the Tanzil Quran transcription corpus\footnote{\url{http://opus.nlpl.eu/}}.

\insertevolution

\subsection{Analysis}

Comparing in-language performance in Table~\ref{tab:results_all}, we observe that, when using BiLSTMs, results are consistently better when we take the dimension-wise maximum over all hidden states (BiLSTM-max) compared to taking the last hidden state (BiLSTM-last). Unsuprisingly, BiLSTM results are better than the pretrained CBOW approach for all languages. As in \citet{bowman2015large}, we also observe the superiority of Bi\-LSTM encoders over CBOW, even when fine-tuning the word embeddings of the latter on the MultiNLI training set, thereby again confirming that the NLI task requires more than just word information. Both of these findings confirm previously published results \cite{conneau2017supervised}.

Table~\ref{tab:results_all} shows that translation offers a strong baseline for XLU. Within translation, {\sc translate test} appears to perform consistently better than {\sc translate train} for all languages. The best cross-lingual results in our evaluation are obtained by the {\sc translate test} approach for all cross-lingual directions. 
Within the translation approaches, as expected, we observe that cross-lingual performance depends on the quality of the translation system. In fact, translation-based results are very well-correlated with the BLEU scores for the translation systems
; XNLI performance for three of the four languages with the best translation systems (comparing absolute BLEU, Table~\ref{tab:bleu}) is above 70\%. 
This performance is still about three points below the English NLI performance of 73.7\%. This slight drop in performance may be related to translation error, changes in style, or artifacts introduced by the machine translation systems that result in discrepancies between the training and test data.


For cross-lingual performance, we observe a healthy gap between the English results and the results obtained on other languages. For instance, for French, we obtain 67.7\% accuracy when classifying French pairs using our English classifier and multilingual sentence encoder. When using our alignment process, our method is competitive with the \textsc{translate train} baseline, suggesting that it might be possible to encode similarity between languages directly in the embedding spaces generated by the encoders. However, these methods are still below the other machine translation baseline \textsc{translate test}, which significantly outperforms the multilingual sentence encoder approach by up to 6\% (Swahili). These production systems have been trained on much larger training data than the ones used for the alignment loss (section ~\ref{SectDataParallel}), which can partly explain the superiority of this method over the baseline. At inference time, the multilingual sentence encoder approach is however much cheaper than the \textsc{translate test} baseline, and this method also does not require any machine translation system. Interestingly, the two points difference in accuracy between X-BiLSTM-last and X-BiLSTM-max is maintained across languages, which suggests that having a stronger encoder in English also positively impacts the transfer results on other languages.

\insertABLATION

For \textsc{x-bilstm} French, Urdu and Arabic encoders, we plot in Figure~\ref{fig:evolution} the evolution of XNLI dev accuracies and the alignment losses during training. The latter are computed using XNLI parallel dev sentences. We observe a strong correlation between the alignment losses and XNLI accuracies. As the alignment on English-Arabic gets better for example, so does the accuracy on XNLI-ar. One way to understand this is to recall that the English classifier takes as input the vector $[u,v,\vert u-v \vert, u*v]$ where $u$ and $v$ are the embeddings of the premise and hypothesis. So this correlation between $\mathcal{L}_{\text{align}}$ and the accuracy suggests that, as English and Arabic embeddings $[u_{\text{en}}, v_{\text{en}}]$ and $[u_{\text{ar}}, v_{\text{ar}}]$ get closer for parallel sentences (in the sense of the L2-norm), the English classifier gets better at understanding Arabic embeddings $[u_{\text{ar}},v_{\text{ar}},\vert u_{\text{ar}}-v_{\text{ar}} \vert, u_{\text{ar}}*v_{\text{ar}}]$ and thus the accuracy improves. We observe some over-fitting for Urdu, which can be explained by the small number of parallel sentences (64k) available for that language.

In Table~\ref{tab:ablation}, we report the validation accuracy using BiLSTM-max on three languages with different training hyper-parameters. Fine-tuning the embeddings does not significantly impact the results, suggesting that the LSTM alone is ensuring alignment of parallel sentence embeddings. We also observe that the negative term is not critical to the performance of the model, but can lead to slight improvement in Chinese (up to 1.6\%).

\section{Conclusion}
A typical problem in industrial applications is the lack of supervised data for languages other than English, and particularly for low-resource languages. Since annotating data in every language is not a realistic approach, there has been a growing interest in cross-lingual understanding and low-resource transfer in multilingual scenarios. In this work, we extend the development and test sets of the Multi-Genre Natural Language Inference Corpus to \nlangs languages, including low-resource languages such as Swahili and Urdu. Our dataset, dubbed XNLI, is designed to address the lack of standardized evaluation protocols in cross-lingual understanding, and will hopefully help the community make further strides in this area. We present several approaches based on cross-lingual sentence encoders and machine translation systems. While machine translation baselines obtained the best results in our experiments, these approaches rely on computationally-intensive translation models either at training or at test time. We found that cross-lingual encoder baselines provide an encouraging and efficient alternative, and that further work is required to match the performance of translation based methods.

\subsubsection*{Acknowledgments}

This project has benefited from financial support to Samuel R. Bowman by Google, Tencent Holdings, and Samsung Research.

\bibliography{emnlp2018}
\bibliographystyle{acl_natbib_nourl}

\end{document}

%% file: content/tables.tex

\newcommand{\insertexamples}{
    \renewcommand{\arraystretch}{1.2}
    \begin{table*}[t!]
    \small
        \resizebox{1\linewidth}{!}{
        \begin{tabular}{m{1cm}m{10.2cm}ll}
        \toprule
        Language
        & Premise / Hypothesis
        & Genre
        & Label\\
        \midrule
        English
            & \begin{tabular}[x]{@{}l@{}}
                You don't have to stay there.\\
                You can leave.
            \end{tabular}
            & Face-To-Face
            & Entailment\\
        \midrule
        French
            & \begin{tabular}[x]{@{}l@{}}
                La figure 4 montre la courbe d'offre des services de partage de travaux.\\
                Les services de partage de travaux ont une offre variable.
            \end{tabular}
            & Government
            & Entailment\\
        \midrule
        Spanish
            & \begin{tabular}[x]{@{}l@{}}
                Y se estremeció con el recuerdo.\\
                El pensamiento sobre el acontecimiento hizo su estremecimiento.
            \end{tabular}
            & Fiction
            & Entailment\\
        \midrule
        German
            & \begin{tabular}[x]{@{}l@{}}
                Während der Depression war es die ärmste Gegend, kurz vor dem Hungertod.\\
                Die Weltwirtschaftskrise dauerte mehr als zehn Jahre an.
            \end{tabular}
            & Travel
            & Neutral\\
        \midrule
        Swahili
            & \begin{tabular}[x]{@{}l@{}}
                Ni silaha ya plastiki ya moja kwa moja inayopiga risasi.\\
                Inadumu zaidi kuliko silaha ya chuma.
            \end{tabular}
            & Telephone
            & Neutral\\
        \midrule
        Russian
            & \begin{tabular}[x]{@{}l@{}}
                \begin{otherlanguage*}{russian}И мы занимаемся этим уже на протяжении 85 лет.\end{otherlanguage*}\\
                \begin{otherlanguage*}{russian}Мы только начали этим заниматься.\end{otherlanguage*}
            \end{tabular}
            & Letters
            & Contradiction\\
        \midrule
        Chinese
            & \begin{tabular}[x]{@{}l@{}}
                \begin{CJK}{UTF8}{gbsn}让我告诉你，美国人最终如何看待你作为独立顾问的表现。\end{CJK}\\
                \begin{CJK}{UTF8}{gbsn}美国人完全不知道您是独立律师。\end{CJK}
            \end{tabular}
            & Slate
            & Contradiction\\
        \midrule
        Arabic
            & 
            \raisebox{-0.1\totalheight}{\includegraphics[scale=0.4]{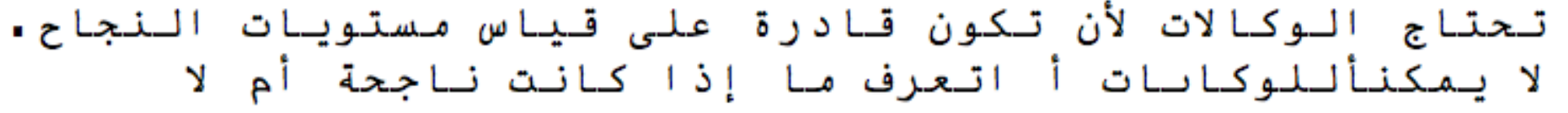}}
            & Nine-Eleven
            & Contradiction\\
        \bottomrule
        \end{tabular}
        }
        \caption{Examples (premise and hypothesis) from various languages and genres from the XNLI corpus.
        \label{tab:examples}}
    \end{table*}
    \renewcommand{\arraystretch}{1}
}

\newcommand{\insertstats}{
  \begin{table*}[t!]
        \begin{center}
            \resizebox{1\linewidth}{!}{
            \begin{tabular}[b]{l|*{15}{@{\,\,}c@{\,\,}}}
                \toprule
                & en & fr & es & de & el & bg & ru & tr & ar & vi & th & zh & hi & sw & ur\\
                \midrule
                Premise   & 21.7 & 24.1 & 22.1 & 21.1 & 21.0  & 20.9 & 19.6 & 16.8 & 20.7 & 27.6 & 22.1 & 21.8 & 23.2 & 18.7 & 24.1\\
                Hypothesis & 10.7 & 12.4 & 10.9 & 10.8 & 10.6 &  10.4 &  9.7 & 8.4  & 10.2 & 13.5 & 10.4 & 10.8 & 11.9 &  9.0 & 12.3\\
                \bottomrule
            \end{tabular}
            }
            \caption{Average number of tokens per sentence in the XNLI corpus for each language.
            \label{tab:stats_all}}
        \end{center}
        \vspace{-0.4cm}
    \end{table*}
}

\newcommand{\insertppmianalysis}{
  \begin{table*}[h]
        \begin{center}
            \begin{tabular}[b]{l|c|c|}
            & word & languages \\
            \hline
            contradiction & never & en, es, fr, de, ar, bg, ru\\
            entailment & some & en, es, ar, bg, ru, zh\\
            neutral & because & en, es, fr, de, bg, hi, ru, sw, zh \\
            \hline
            \end{tabular}
            \caption{\textrm{Hypothesis words indicative of class. Most common word in the top two PMI(word, class) across languages for each class, and the languages in which it came up} 
            \label{tab:ppmianalysis}}
        \end{center}
    \end{table*}
}

\newcommand{\insertppmianalysisV}{
  \begin{table}[h!]
        \begin{center}
            \begin{tabular}[b]{l|*{3}{p{1.7cm}|}}
            & contradictory & entailment & neutral \\
            \hline
            word & never  & some & because\\
            \hline
            langs
             & en, es, fr, de, ar, bg, ru
             & en, es, ar, bg, ru, zh
             & en, es, fr, de, bg, hi, ru, sw, zh \\
            \hline
            \end{tabular}
            \caption{\textrm{Hypothesis words indicative of class. Most common word in the top two PMI(word, class) across languages for each class, and the languages in which it came up.} 
            \label{tab:ppmianalysis}}
        \end{center}
    \end{table}
}

\newcommand{\insertppmi}{
  \begin{table*}[h]
        \begin{center}
            \begin{tabular}[b]{l|c|c|c|}
            & contradiction & entailment & neutral  \\
            \hline
            en & no (15) & some (8) & because (9) \\
            fr & n' (33) & chose (6) & DIGITS (15) \\
            es & nunca (12) & algunos (6) & DIGITS (15) \\
            de & keine (14) & einen (8) & weil (8) \\
            bg & никога (12) & някои (7) & защото (8) \\
            ru & никогда (11) & есть (8) & DIGITS (15) \\
            ar & word (0.00) & word (0.00) & word (0.00) \\
            hi & word (0.00) & word (0.00) & word (0.00) \\
            sw & word (0.00) & word (0.00) & word (0.00) \\
            ur & word (0.00) & word (0.00) & word (0.00) \\
            \hline
            \end{tabular}
            \caption{\textbf{TBD} 
            \label{tab:stats_all}}
        \end{center}
    \end{table*}
}

\newcommand{\insertXNLIresults}{
    \begin{table*}[t]
        \begin{center}
            \resizebox{1\linewidth}{!}{
            \begin{tabular}[b]{l|ccccccccccccccc}
            \toprule
            & en & fr & es & de & el & bg & ru & tr & ar & vi & th & zh & hi & sw & ur \\

                \midrule
                \multicolumn{15}{l}{\it Machine translation baselines (\textsc{translate train})} \\
                \midrule
                BiLSTM-last & 71.0 & 66.7 & 67.0 & 65.7 & 65.3 & 65.6 & 65.1 & 61.9 & 63.9 & 63.1 & 61.3 & 65.7 & 61.3 & 55.2 & 55.2\\ 
                BiLSTM-max & \bf73.7 & 68.3 & 68.8 & 66.5 & 66.4 & 67.4 & 66.5 & 64.5 & 65.8 & 66.0 & 62.8 & 67.0 & 62.1 & 58.2 & 56.6 \\
                \midrule
                \multicolumn{15}{l}{\it Machine translation baselines (\textsc{translate test})} \\
                \midrule
                BiLSTM-last & 71.0 & 68.3 & 68.7 & 66.9 & 67.3 & 68.1 & 66.2 & 64.9 & 65.8 & 64.3 & 63.2 & 66.5 & 61.8 & 60.1 & 58.1 \\                BiLSTM-max & \bf73.7 & \bf 70.4 & {\bf 70.7} & \bf68.7 & \bf69.1 & {\bf 70.4} & \bf67.8 & \bf66.3 & \bf66.8 & \bf66.5 & \bf64.4 & \bf68.3 & \bf64.2 & \bf61.8 & \bf59.3 \\

                \midrule
                \multicolumn{15}{l}{\it Evaluation of XNLI multilingual sentence encoders (in-domain)} \\ 
                \midrule
                X-BiLSTM-last & 71.0 & 65.2 & 67.8 & 66.6 & 66.3 & 65.7 & 63.7 & 64.2 & 62.7 & 65.6 & 62.7 & 63.7 & 62.8 & 54.1 & 56.4 \\
                X-BiLSTM-max & \bf73.7 & 67.7 & 68.7 & 67.7 & 68.9 & 67.9 & 65.4 & 64.2 & 64.8 & 66.4 & 64.1 & 65.8 & 64.1 & 55.7 & 58.4 \\

                \midrule
                \multicolumn{15}{l}{\it Evaluation of pretrained multilingual sentence encoders (transfer learning)} \\ 
                \midrule
                X-CBOW & 64.5 & 60.3 & 60.7 & 61.0 & 60.5 & 60.4 & 57.8 & 58.7 & 57.5 & 58.8 & 56.9 & 58.8 & 56.3 & 50.4 & 52.2 \\
    
                \bottomrule
            \end{tabular}
            }
            \caption{Cross-lingual natural language inference (XNLI) test accuracy for the \nlangs languages.
            \label{tab:results_all}}
        \end{center}
       \vspace{-0.4cm}
    \end{table*}
}

\newcommand{\insertABLATION}{
    \begin{table}[h!]
        \begin{center}
            \small
            \begin{tabular}[b]{l|ccc}
            \toprule
            & fr & ru & zh \\
            \midrule
            $ft=1, \lambda=0.25$ [default] & 68.9 & 66.4 & 67.9 \\
            $ft=1, \lambda=0.0$ (no negatives) & 67.8 & 66.2 & 66.3 \\
            $ft=1, \lambda=0.5$& 64.5 & 61.3 & 63.7 \\
            $ft=0, \lambda=0.25$ & 68.5 & 66.3 & 67.7 \\
            \bottomrule
            \end{tabular}
            \caption{\textbf{Validation accuracy using BiLSTM-max.} Default setting corresponds to $\lambda=0.25$ (importance of the negative terms) and uses fine-tuning of the target lookup table ($ft=$1).
            \label{tab:ablation}}
        \end{center}
    \end{table}
}

\newcommand{\insertevolution}{
    \begin{table}[h]
        \includegraphics[scale=0.3]{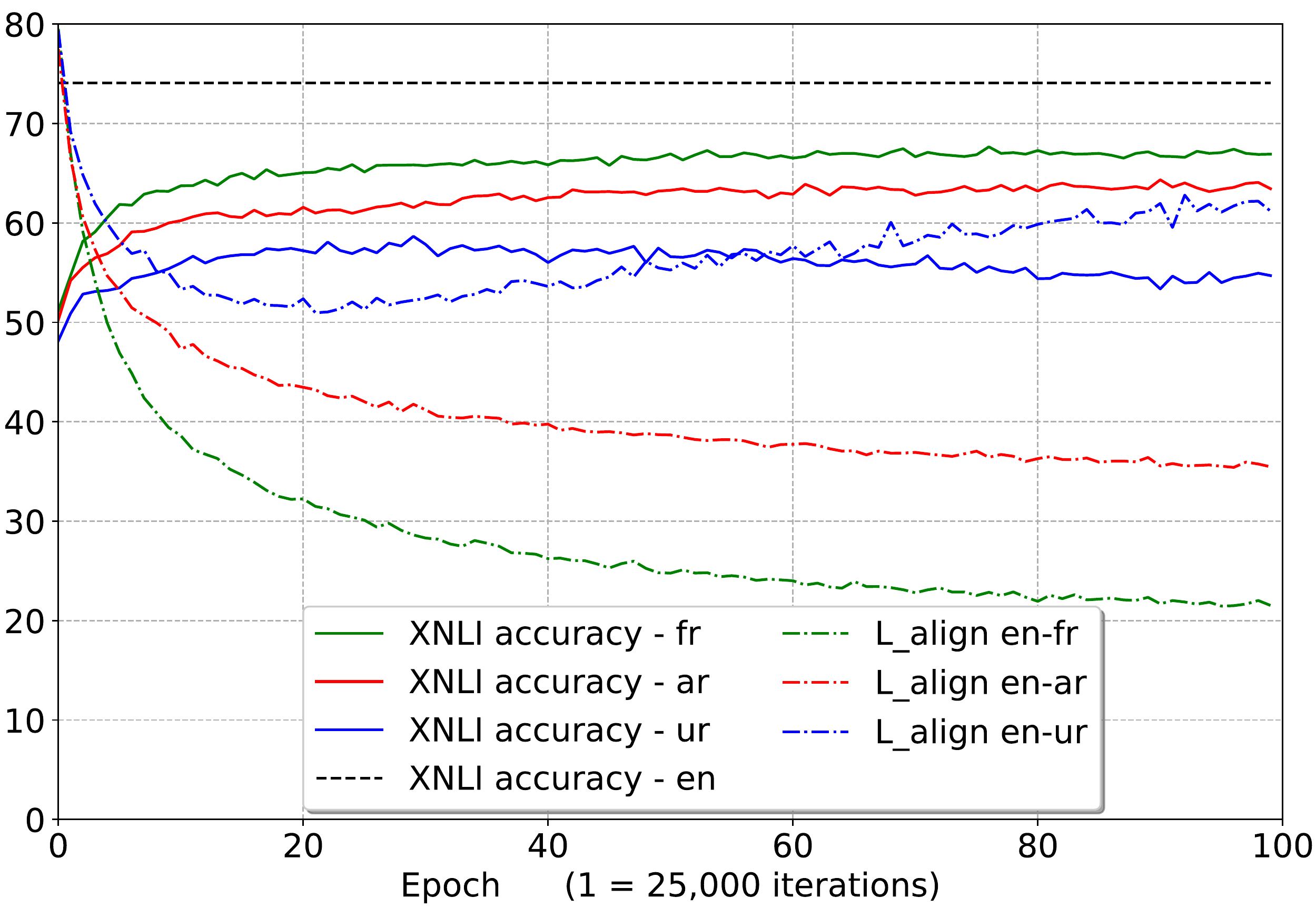}
        \captionof{figure}{Evolution along training of alignment losses and \textsc{x-bilstm} XNLI French (fr), Arabic (ar) and Urdu (ur) accuracies. Observe the correlation between $\mathcal{L}_{\text{align}}$ and accuracy.
        \label{fig:evolution}}
    \vspace{-0.4cm}
    \end{table}
}

\newcommand{\insertalignment}{
\begin{center}
    \begin{table*}[t!]
        \includegraphics[scale=0.6]{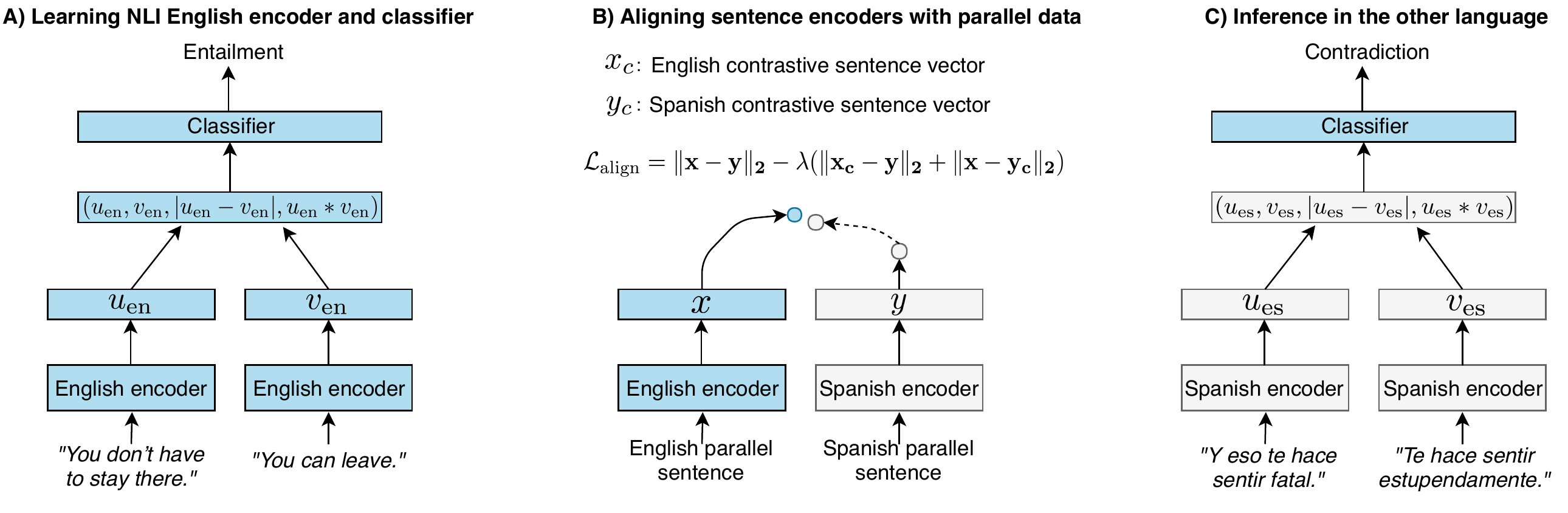}
        \captionof{figure}{\textbf{Illustration of language adaptation by sentence embedding alignment.} A) The English encoder and classifier in blue are learned on English (\textit{in-domain}) NLI data. The encoder can also be pretrained (\textit{transfer learning}). B) The Spanish encoder in gray is trained to mimic the English encoder using parallel data. C) After alignment of the encoders, the classifier can make predictions for Spanish.
        \label{fig:alignment}}
    \end{table*}
\end{center}
}




\newcommand{\insertwordtranslationtable}{
    \begin{table*}[h]
    \begin{center}
    \resizebox{1\linewidth}{!}{
    \begin{tabular}[b]{l|*{14}{@{\,\,}c@{\,\,}}}
    \toprule
    & fr & es & de & el & bg & ru & tr & ar & vi & th & zh & hi & sw & ur\\
    \midrule
    XX-En BLEU & 41.2 & 45.8 & 39.3 & 42.1 & 38.7 & 27.1 & 29.9  & 35.2 & 23.6 & 22.6 & 24.6 & 27.3 & 21.3 & 24.4 \\

    En-XX BLEU & 49.3 & 48.5 & 38.8 & 42.4 & 34.2 & 24.9 & 21.9 & 15.8 & 39.9 & 21.4 & 23.2 & 37.5 & 24.6 & 24.1 \\
    Word translation P@1 & 73.7 & 73.9 & 65.9 & 61.1 & 61.9 & 60.6 & 55.0 & 51.9 & 35.8 & 25.4 & 48.6 & 48.2 & - & - \\
    \bottomrule
    \end{tabular}
    }
    \caption{BLEU scores of our translation models (XX-En) P@1 for multilingual word embeddings.}
    \label{tab:bleu}
    \end{center}
    \end{table*}

}

\newcommand{\insertwordtranslationtableOld}{
    \begin{table*}[h!]
    \begin{center}
    \small
    \begin{tabular}[b]{|l|c|c|c|c|c|c|c|c|c|c|}
    \hline
    & ar - en & bg - en & de - en & es - en & fr - en & hi - en & ru - en & sw - en & ur - en & zh - en\\
    \hline
    Accuracy - NN   & 51.93 & 61.94 & 65.87 & 73.85 & 73.67 & 48.24 & 60.62 & - & - & 48.61\\
    \hline
    \end{tabular}
    \caption{Word translation accuracy (with precision at 1) on the word translation task. \red{it may be better to use the same order of the languages than in Tab~3}}
    \label{tab:wordtranslation}
    \end{center}
    \end{table*}
}

\newcommand{\insertXNLIresultsShort}{
    \begin{table*}[h!]
        \begin{center}
            \resizebox{1\linewidth}{!}{
            \begin{tabular}[b]{l|ccccccccccccccc}
            \toprule
            & en & fr & es & de & el & bg & ru & tr & ar & vi & th & zh & hi & sw & ur \\
                \midrule
               translate train & \bf73.7 & 68.3 & 68.8 & 66.5 & 66.4 & 67.4 & 66.5 & 64.5 & 65.8 & 66.0 & 62.8 & 67.0 & 62.1 & 58.2 & 56.6 \\
                \midrule
                translate test & \bf73.7 & \bf 70.4 & {\bf 70.7} & \bf68.7 & \bf69.1 & {\bf 70.4} & \bf67.8 & \bf66.3 & \bf66.8 & \bf66.5 & \bf64.4 & \bf68.3 & \bf64.2 & \bf61.8 & \bf59.3 \\
                \midrule
                XNLI multilingual sentence encoders & \bf73.7 & 67.7 & 68.7 & 67.7 & 68.9 & 67.9 & 65.4 & 64.2 & 64.8 & 66.4 & 64.1 & 65.8 & 64.1 & 55.7 & 58.4 \\

                \midrule
                XLU CBOW word embeddings  & 64.5 & 60.3 & 60.7 & 61.0 & 60.5 & 60.4 & 57.8 & 58.7 & 57.5 & 58.8 & 56.9 & 58.8 & 56.3 & 50.4 & 52.2 \\
    
                \bottomrule
            \end{tabular}
            }
            \caption{Results
            \label{tab:results_short}}
        \end{center}
    \end{table*}
}